\documentclass[conference]{IEEEtran}
\usepackage{cite}

\ifCLASSINFOpdf
   \usepackage[pdftex]{graphicx}
\else
\fi
\usepackage[cmex10]{amsmath}
\usepackage{amsfonts}
\usepackage{bm}
\usepackage{xfrac}
\usepackage{mathtools}

\usepackage[linesnumbered, vlined, ruled, commentsnumbered]{algorithm2e}

\usepackage{array}
\usepackage{multicol}
\usepackage{colortbl}

\usepackage{multirow}

\ifCLASSOPTIONcompsoc
  \usepackage[caption=false,font=normalsize,labelfont=sf,textfont=sf]{subfig}
\else
  \usepackage[caption=false,font=footnotesize]{subfig}
\fi
\usepackage{ragged2e}

\usepackage{fixltx2e}
\usepackage{balance}
 \usepackage{dblfloatfix}

\usepackage{url}
\usepackage[colorlinks=true, citecolor=green, linkcolor=red, urlcolor=blue]{hyperref}
\usepackage[hyperpageref]{backref}
\usepackage{lipsum}
\usepackage{graphicx}  
\usepackage[export]{adjustbox}

\usepackage[caption=false,font=footnotesize]{subfig}  

\usepackage{longtable}
\usepackage{booktabs}
\usepackage{ifthen}

\begin{document}
\bstctlcite{IEEEexample:BSTcontrol}
%
\title{CellGenNet: A Knowledge-Distilled Framework for Robust Cell Segmentation in Cancer Tissues}


\author{\IEEEauthorblockN{Srijan Ray$^{a}$, Bikesh K. Nirala$^{b}$, Jason T. Yustein$^{b}$, Sundaresh Ram$^{c, d, e}$}
\IEEEauthorblockA{$^{a}$College of Computing, Georgia Institute of Technology, Atlanta, GA, USA\\$^{b}$Department of Pediatrics, Emory University, Atlanta, GA, USA\\ $^{c}$Departments of Radiology \& Imaging Sciences, Emory University, Atlanta, GA, USA\\ $^{d}$Department of Biomedical Engineering, Emory University \& Georgia Institute of Technology, Atlanta, GA, USA\\ $^{e}$School of Electrical and Computer Engineering, Georgia Institute of Technology, Atlanta, GA, USA\\
Email: \{{\tt \href{mailto:sray99@gatech.edu}{sray99},\href{mailto:sram3@gatech.edu}{sram3}}\}{\tt \href{mailto:sray99@gatech.edu}{@gatech.edu}}}
}


%


\maketitle

\begin{abstract}
Accurate nuclei segmentation in microscopy whole slide images (WSIs) remains challenging due to variability in staining, imaging conditions, and tissue morphology. We propose \textit{CellGenNet}, a knowledge distillation framework for robust cross-tissue cell segmentation under limited supervision. \textit{CellGenNet} adopts a student-teacher architecture, where a capacity teacher is trained on sparse annotations and generates soft pseudo-labels for unlabeled regions. The student is optimized using a joint objective that integrates ground-truth labels, teacher-derived probabilistic targets, and a hybrid loss function combining binary cross-entropy and Tversky loss, enabling asymmetric penalties to mitigate class imbalance and better preserve minority nuclear structures. Consistency regularization and layerwise dropout further stabilize feature representations and promote reliable feature transfer. Experiments across diverse cancer tissue WSIs show that \textit{CellGenNet} improves segmentation accuracy and generalization over supervised and semi-supervised baselines, supporting scalable and reproducible histopathology analysis.
\end{abstract}



%
\IEEEpeerreviewmaketitle

\section{Introduction}
Accurate nuclei segmentation in microscopy whole slide images (WSIs) is critical for quantitative imaging analyses, enabling extraction of morphometric descriptors, spatial organization metrics, and phenotype-specific biomarkers at single-cell resolution \cite{ram12size, ram2017sparse, ram13symmetry}. These measurements underpin downstream applications such as tumor microenvironment characterization, tissue architecture modeling, and computational diagnostic grading \cite{kumar2017dataset, ram2020combined, kumar2019multi}. Manual annotation is impractical for large-scale WSIs due to high spatial resolution, extensive field of view, extreme nuclei density, and inter-observer variability. Segmentation is further challenged by heterogeneous staining, scanner-dependent illumination, out-of-focus regions, folded tissue, and substantial variability in nuclear morphology \cite{graham2019hover,lin2023nuclei,nunes2025survey}. As digital pathology datasets grow increasingly heterogeneous, automated segmentation methods must remain robust and generalizable across diverse imaging conditions and tissue types.

Early nuclei segmentation approaches relied on classical image-processing or model-based techniques, including multi-scale filtering \cite{ram12size}, radial symmetry transforms \cite{ram2016size}, watershed algorithms, and graph-cut optimization \cite{nandy2015segmentation}. While effective in controlled settings, these methods assume consistent nuclear morphology or intensity distributions, limiting their generalization. High computational costs, particularly for 3D or high-resolution WSIs, further constrain scalability in high-throughput workflows.

Deep learning has recently enabled more generalizable solutions. Models such as \textit{StarDist} \cite{Schmidt_2018, weigert2022}, \textit{Cellpose} \cite{Stringer2021Cellpose}, and \textit{InstanSeg} \cite{goldsborough2024instansegembeddingbasedinstancesegmentation} achieve robust segmentation across many tissue types. \textit{StarDist} represents objects as star-convex polygons by predicting center probabilities and radial distances, \textit{Cellpose} employs a flow-based approach to capture diverse morphologies, and \textit{InstanSeg} extends a modified U-Net backbone \cite{ronneberger2015u} by clustering pixel embeddings around optimally selected seeds. Despite these advances, these methods often fail on highly irregular, elongated, or non-convex nuclei, such as those found in heterogeneous tissues containing macrophages, dendritic cells, eosinophils, and other rare cell types. Errors in segmentation can lead to incomplete results and require additional manual correction, limiting practical efficiency \cite{guo2025assessmentcellnucleiai}.

Task-specific networks, typically U-Net or ResNet-based, trained on manually annotated nuclei masks, can achieve higher accuracy \cite{LAL2021104075, reviewOfNucleiDetectionSegmentationMethods}. However, these specialized models require large annotated datasets—often hundreds of thousands of nuclei—to capture morphological variability. Generating such datasets reintroduces the bottleneck that automated segmentation aims to eliminate.

To overcome these challenges, we propose \textit{CellGenNet}, a robust knowledge-distillation framework for accurate and generalizable nuclei segmentation that does not require large manually annotated datasets. Unlike prior methods that either rely on generalized models with limited morphological sensitivity or require labor-intensive expert annotations, \textit{CellGenNet} distills structural knowledge from a \textit{StarDist} teacher into a lightweight U-Net student trained on pseudo–ground truth masks. Our contributions are threefold: (1) a hybrid loss combining Binary Cross-Entropy with a Tversky loss calibrated for missing or incomplete nuclei, enabling preservation of small or irregular nuclear structures often absent in pseudo labels; (2) consistency regularization via custom horizontal and vertical split-and-flip augmentations, together with layerwise dropout, to stabilize feature representations and improve robustness to heterogeneous nuclear shapes, densities, and staining variations; and (3) demonstration that even with a simple network architecture, \textit{CellGenNet} outperforms its teacher and widely used segmentation frameworks---including \textit{StarDist} \cite{Schmidt_2018, weigert2022}, \textit{Cellpose} \cite{Stringer2021Cellpose}, and \textit{InstanSeg} \cite{goldsborough2024instansegembeddingbasedinstancesegmentation}---particularly on irregular, elongated, or densely clustered nuclei. We validate \textit{CellGenNet} on both the open NuInsSeg dataset \cite{Mahbod2024} and our proprietary Osteosarcoma dataset, showing consistent improvements in accuracy, generalizability, and morphological completeness.

\section{Methods}
Our framework, CellGenNet, employs a semi-supervised student–teacher knowledge distillation (KD) \cite{hinton2015distillingknowledgeneuralnetwork} approach to train a robust and generalizable nuclei segmentation model. The student network learns from both sparse ground-truth annotations and pseudo-labels generated by a high-capacity teacher. Below, we describe the network architecture, training procedure, and loss functions used in CellGenNet.

\subsection{CellGenNet Architecture}

CellGenNet is built on a student–teacher knowledge distillation (KD) framework designed to function with limited or no ground-truth labels. A pre-trained \textit{StarDist} model serves as the teacher, while the student network is based on a U-Net architecture \cite{ronneberger2015u}. The U-Net’s contracting and expanding paths capture contextual information around nuclei, providing a rich object representation. The network employs four downsampling and upsampling steps with ReLU activations throughout. In the contraction path, each layer consists of two 3 × 3 convolutions followed by 2 × 2 max pooling (stride 1 in each dimension). The expansion path performs 2 × 2 upsampling, concatenates cropped feature maps from the corresponding contraction layer, and applies two 3 × 3 convolutions with ReLU. The final layer uses a 1 × 1 convolution to produce a binary segmentation map. The full network comprises 18 convolutional layers, and input patches are zero-padded to ensure the output matches the input dimensions. 

\begin{figure}[!t]
\includegraphics[width= 3.4in, height=1.7in]{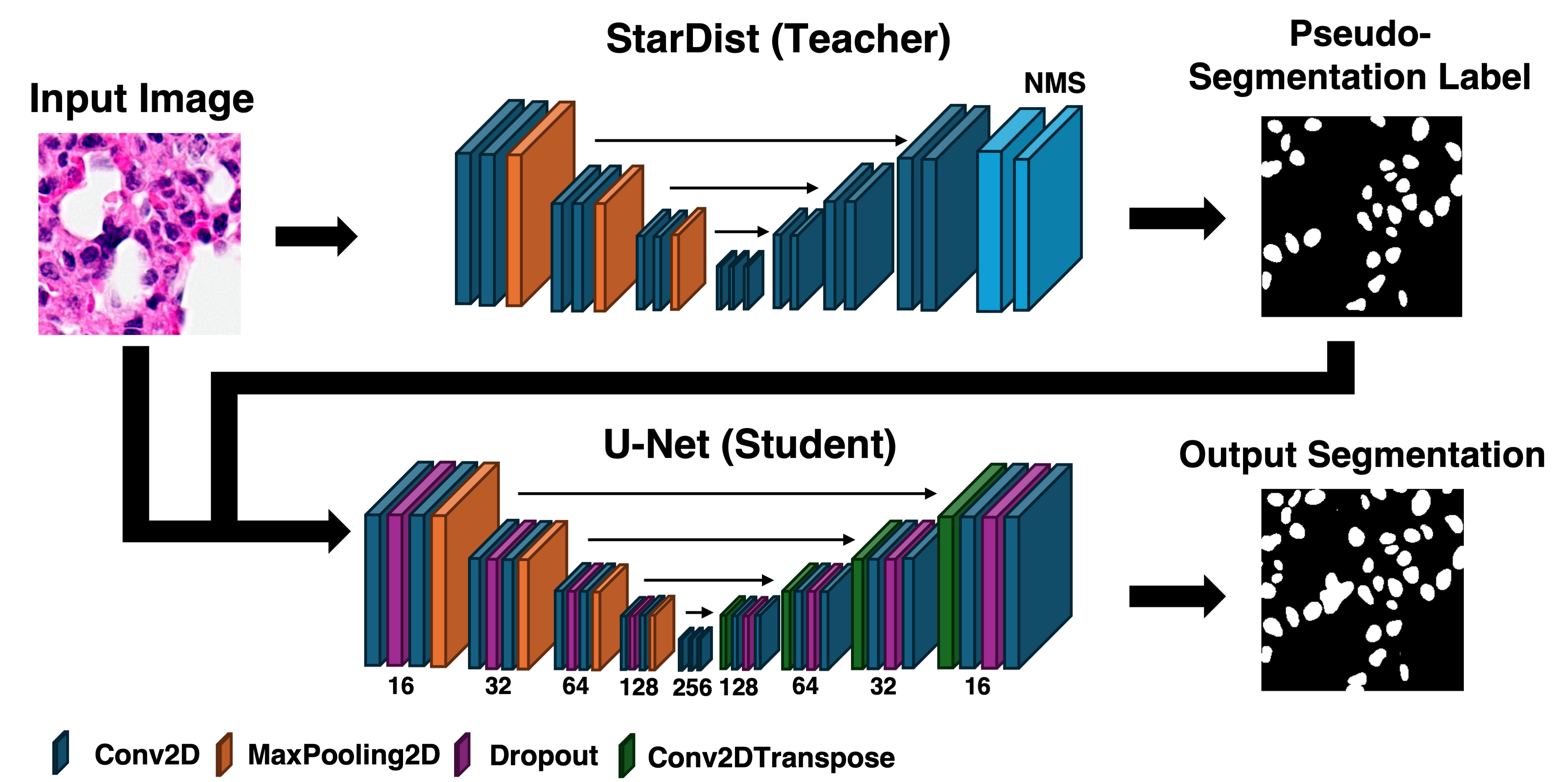}
\caption{An overview of \textit{CellGenNet} student–teacher segmentation framework.}
\label{fig1}
\vspace{-5mm}
\end{figure}

Training follows a semi-supervised approach. First, the entire training set is processed through the \textit{StarDist} teacher model to generate pseudo–ground truth labels. The U-Net student is then trained from scratch using the original image patches along with these pseudo labels. A 10\% validation subset, also labeled by \textit{StarDist}, is used to monitor overfitting to the pseudo labels and guide hyperparameter selection. The resulting trained student network constitutes the final \textit{CellGenNet} model. The entire network details are illustrated in Fig.~\ref{fig1}.

\subsection{Bias-Correcting Compound Loss Function}
\label{loss}
To account for missing or incomplete nuclei in the pseudo–ground truth labels generated by the teacher, the student network is trained using a compound loss function:

\begin{equation}
\label{eq1}
    \mathcal{L}_{\text{total}} = 0.4*\mathcal{L}_{\text{BCE}} + 0.6*\mathcal{L}_{\text{Tversky}}
\end{equation}

where $\mathcal{L}_{\text{BCE}}$ is defined as:
\begin{equation}
\label{eq2}
    \mathcal{L}_{\text{BCE}} = - \frac{1}{N} \sum_{i=1}^{N} \left[x_{i}\log(\hat{x_{i}}) + (1 - x_{i})\log(1-\hat{x_{i}})\right]
\end{equation}
Here, $x_{i}$ and $\hat{x_{i}}$ are the ground truth and predicted values for pixel $i$ and $N$ is the total number of pixels. 
The Tversky loss ($\mathcal{L}_{\text{Tversky}}$) is defined as:
\begin{equation}
\label{eq3}
    \mathcal{L}_{\text{Tversky}} = \frac{TP}{TP+\alpha FP + \beta FN}
\end{equation}

where $TP$, $FP$, and $FN$ denote true positives, false positives, and false negatives, respectively. The $\alpha$ and $\beta$ parameters allow asymmetric weighting, emphasizing false negatives over false positives. This encourages the student to learn from nuclei correctly captured by the teacher while also recovering nuclei that may be missing from pseudo labels, without heavily penalizing potential “false positives.”

Robustness is further enhanced through consistency regularization using custom horizontal and vertical split-and-flip augmentations, encouraging the network to produce consistent predictions across transformed views of the same patch. Layerwise dropout is also applied to stabilize feature learning and prevent overfitting. Together, these strategies improve segmentation of irregular, elongated, and densely clustered nuclei across heterogeneous tissue types.

\subsection{Implementation and Training Details}

The student U-Net in \textit{CellGenNet} is trained end-to-end using stochastic gradient descent on the compound loss described in Section~\ref{loss}. Training is performed for 25 epochs, with each epoch consisting of four mini-batches. We optimize the network using RMSProp, which updates weights using exponentially decaying averages of past squared gradients. Bias terms are initialized to zero, and the learning rate is set to $\gamma = 0.001$. RMSProp hyperparameters include an exponential decay rate of $\eta = 1$, a discounting factor of $\rho = 0.9$, and a small constant $\epsilon = 1\times10^{-7}$ to avoid division by zero. Each convolutional layer is followed by a ReLU activation to introduce nonlinearity and stabilize gradient flow. To improve robustness, we apply layerwise dropout with a progressive schedule: dropout begins at 0.1 in the first layer, increases in increments of 0.1 toward the bottleneck, and symmetrically decreases during the upsampling path. For the Tversky loss in equation (\ref{eq3}), we set $\alpha = 0.2$ and $\beta = 0.8$, weighting false negatives more heavily to encourage recovery of nuclei missed in the pseudo labels. The entire \textit{CellGenNet} model is trained from scratch with randomly initialized weights and evaluated on an NVIDIA RTX 4060 GPU (8 GB VRAM).

 \begin{table*}[!b]
 \vspace{-5mm}
 \caption{\sc{Mean Segmentation Performance Measurement (And Standard Deviation) Results}}
 \label{table1}
   \begin{center}
 \vspace{-5mm}
 	\renewcommand{\arraystretch}{1.1}
    \begin{tabular}{>{\centering} m{3.102cm} >{\centering} m{3.102cm} >{\centering}m{1.551cm} >{\centering} m{1.551cm} >{\centering} m{1.551cm} >{\centering} m{1.551cm} >{\centering} m{1.994cm}}
    \hline
 	\rowcolor[gray] {0.8}\textbf{Data} & \textbf{Method} &  \textbf{Dice} & \textbf{IoU} & \textbf{FPR}\\$(\times10^{-1})$ & \textbf{TPR} & \textbf{HD} \tabularnewline \hline
     \multirow{5}{*}{\raisebox{+2em}{\shortstack{\textbf{Osteosarcoma Dataset}\\ 20 image patches}}}
    & Cellpose & 0.47 (0.25) & 0.34 (0.22) & 0.42 (0.86) & 0.37 (0.20) & 66.91 (33.31) \tabularnewline
    & InstanSeg & 0.55 (0.09) & 0.38 (0.09) & \textbf{0.01} (0.01) & 0.39 (0.09) & 59.26 (31.20)  \tabularnewline
    & StarDist & 0.75 (0.05) & 0.60 (0.06) & 0.03 (0.03) & 0.63 (0.06) & 69.69 (40.49) \tabularnewline
    & CellGenNet & \textbf{0.83} (0.07) & \textbf{0.71} (0.09) & 0.05 (0.05) & \textbf{0.77} (0.06) & \textbf{20.34 (10.77)}  \tabularnewline
    \hline
     \multirow{5}{*}{\raisebox{+2em}{\shortstack{\textbf{NuInsSeg}\\123 image patches}}}
    & Cellpose & 0.50 (0.28) & 0.38 (0.25) & 1.64 (2.26) & 0.60 (0.32) & 156.05 (114.70) \tabularnewline
    & InstanSeg & 0.52 (0.18) & 0.37 (0.16) & 0.11 (0.12) & 0.41 (0.19) & 139.82 (77.80)  \tabularnewline
    & StarDist & 0.47 (0.22) & 0.34 (0.18) & \textbf{0.10 (0.16)} & 0.40 (0.24) & 182.10 (110.36) \tabularnewline
    & CellGenNet & \textbf{0.60} (\textbf{0.20}) & \textbf{0.45} (\textbf{0.19}) & 0.50 (0.59) & \textbf{0.63} (\textbf{0.25}) & \textbf{126.18 (81.97)}  \tabularnewline
            \hline
 	\end{tabular}
    \end{center}
 \vspace{-6mm}
 \end{table*}
 
\subsection{Datasets}

We used two distinct datasets to train and evaluate CellGenNet, resulting in two independently trained, dataset-specific models.

\subsubsection{\textbf{Osteosarcoma (Internal) Dataset}}
This internal dataset consists of 8,000 H\&E-stained $256\times256$ image patches extracted from whole-slide images (WSIs) of two tissue regions: the primary tumor site (bone marrow) and the metastatic site (lungs). Each region contributes 4,000 patches, split evenly between full tissue coverage patches—with minimal white background—and partial tissue coverage patches, which contain substantial background. This design reduces the risk of the model erroneously labeling background as nuclei. The dataset was divided into 87.5\% training (7,000 patches), 12.5\% validation (1,000 patches), and a test set consisting of 20 held-out patches manually annotated by an expert.

\subsubsection{\textbf{NuInsSeg Dataset}}
To benchmark performance on a public dataset, we evaluated CellGenNet using the NuInsSeg dataset \cite{Mahbod2024}, which contains 621 H\&E-stained image patches of size $512\times512$ from human and mouse WSIs spanning 20 diverse organs including the brain, heart, bladder, and lung and a wide range of nuclear morphologies. We followed a standard split of 72\% for training (448 images), 8\% for validation (50 images), and 20\% for testing (123 images).




\section{Experiments and Results}

We evaluated the performance of \textit{CellGenNet} by comparing it against three baseline methods—\textit{Cellpose} \cite{Stringer2021Cellpose}, \textit{InstanSeg} \cite{goldsborough2024instansegembeddingbasedinstancesegmentation}, and the teacher model, \textit{StarDist} \cite{Schmidt_2018, weigert2022}—using the two held-out test sets: the 20-image expert-annotated subset from our Osteosarcoma dataset and the 123-image test set from the NuInsSeg benchmark \cite{Mahbod2024}. Given the broad variation in tissue type, species, staining characteristics, and nuclear morphology within NuInsSeg, we expect all models to exhibit lower performance on this dataset compared with the more homogeneous Osteosarcoma dataset.

\subsection{Performance Evaluation}

We evaluated all models using standard segmentation metrics, including Dice similarity coefficient, Intersection over Union (IoU, also known as the Jaccard Index) [1], False Positive Rate (FPR), True Positive Rate (TPR, equivalent to recall or sensitivity), Hausdorff Distance (HD) [5], and the $F_{1}$-score. Let $TP$, $FP$, $FN$, and $TN$ denote the numbers of true-positive, false-positive, false-negative, and true-negative voxels, respectively.

The Dice coefficient measures overlap between predicted and ground-truth segmentations \cite{ram2017sparse}, TPR (recall, sensitivity) = $TP/(TP + FN)$ and FPR ($1-$specificity) = $FP/(FP+TN)$.The IoU (Jaccard Index) quantifies region-level agreement and is computed as IoU = $TP/(TP+FP+FN)$.
To capture boundary-level differences, we compute the Hausdorff Distance (HD), which measures the maximum Euclidean distance from any boundary voxel of one segmentation to the nearest boundary voxel in the other \cite{ram2020combined}. The F1-score, defined as the harmonic mean of precision and recall \cite{ram2016size}, provides an additional coverage-based assessment.

Together, these metrics evaluate pixel-level accuracy, and boundary consistency. An ideal segmentation model maximizes Dice, IoU, TPR, and $F_{1}$-score, while minimizing FPR and HD.

Table~\ref{table1} summarizes segmentation performance on the Osteosarcoma and NuInsSeg test sets, with the highest values for each metric highlighted in bold. On the Osteosarcoma dataset, \textit{CellGenNet} outperforms \textit{Cellpose} by 36\% (Dice) and 37\% (IoU), surpasses \textit{InstanSeg} by 28\% and 33\%, and exceeds \textit{StarDist} by 8\% and 11\%, respectively. It also achieves the highest TPR and lowest HD, indicating fewer missed nuclei and more accurate boundaries. The only metric where it does not lead is FPR, slightly elevated due to the Tversky loss ($\beta$) encouraging over-segmentation to recover nuclei potentially missing in pseudo labels. On the NuInsSeg dataset, CellGenNet similarly attains the highest Dice, IoU, and TPR, reflecting improved coverage across diverse tissues and staining conditions, with the lowest HD and slightly elevated FPR for the same reason.
\begin{figure}[!t]
    \centering
    \subfloat[Comparison of segmentation models on Osteosarcoma nuclei.]
             {\includegraphics[width=0.98\linewidth]{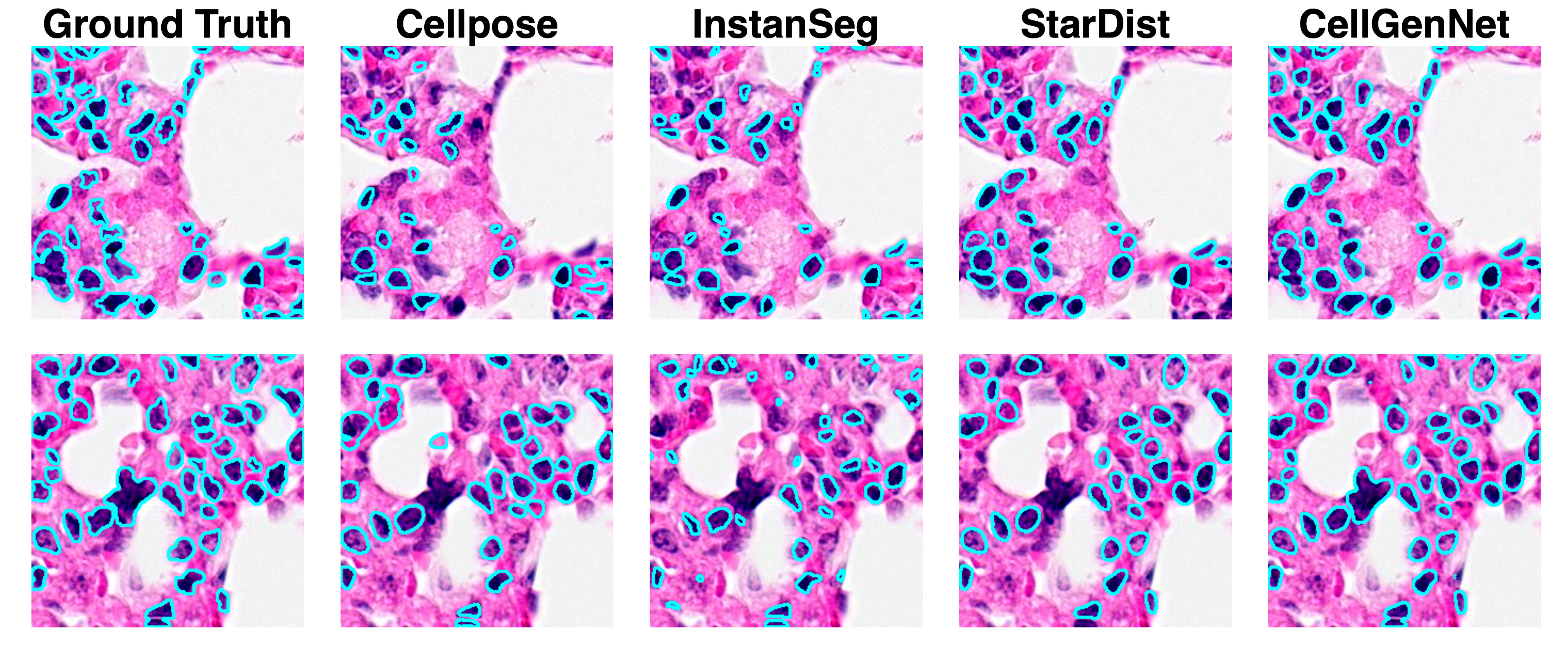}
             \label{fig2:a}}
             
\vspace{0mm}
    
    \subfloat[Comparison of segmentation models on various tissue types (Placenta, Lung, Pancreas) from the NuInsSeg dataset \cite{Mahbod2024}.]
             {\includegraphics[width=0.98\linewidth]{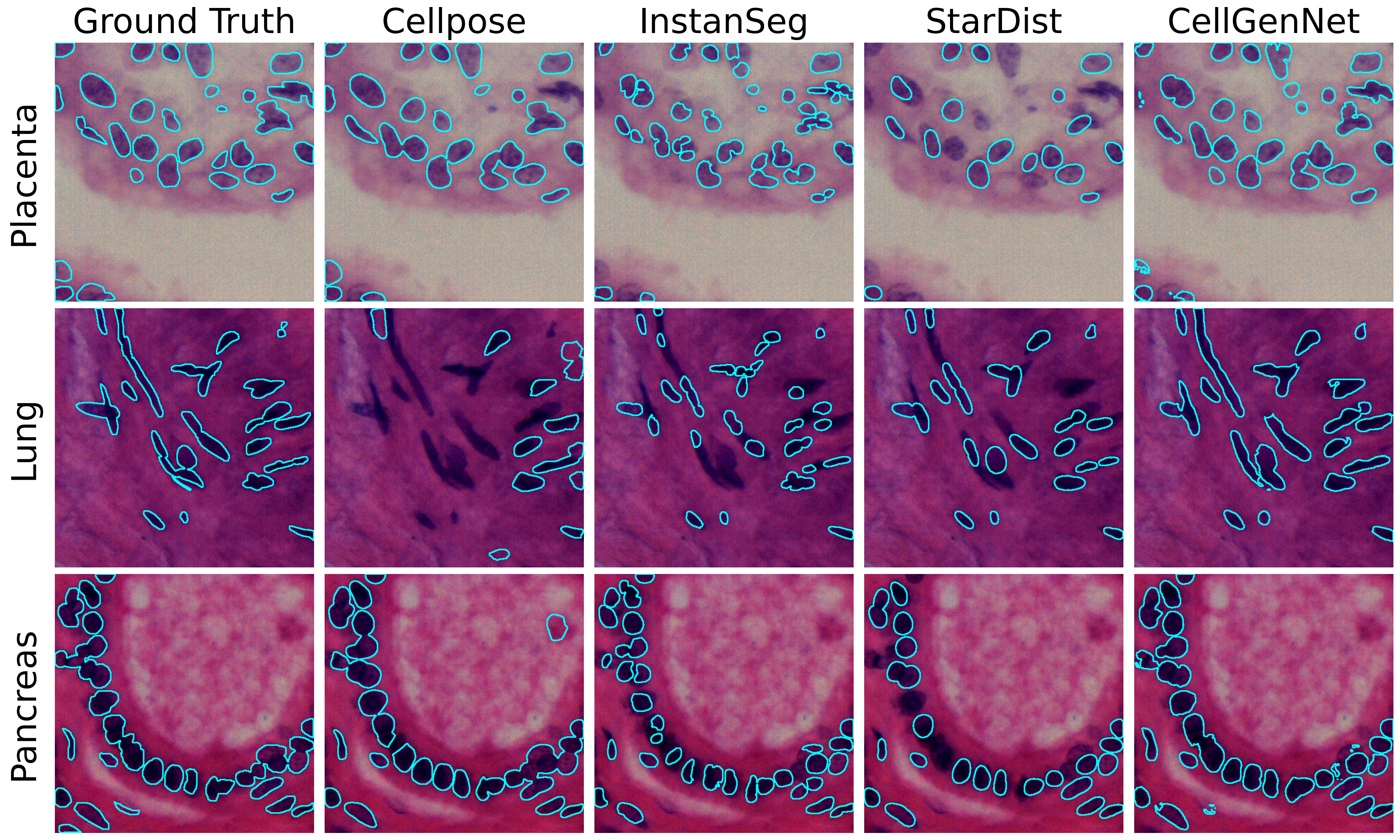}
             \label{fig2:b}}
    \vspace{1mm}
    \caption{Qualitative results of the compared segmentation model outputs against the ground truth for the Osteosarcoma and NuInsSeg Datasets. The segmentation boundary are overlaid onto the images (blue color). Best viewed in zoomed mode.}
    \label{fig2}
      \vspace{1mm}
\end{figure}

Figure~\ref{fig2} shows qualitative segmentation results for both datasets. In Fig.~\ref{fig2:a} (second row), all baseline models fail to capture an irregular cluster of nuclei, whereas CellGenNet segments it accurately. In Fig.~\ref{fig2:b} (“Lung” row), baseline methods struggle with elongated, irregularly shaped nuclei, while CellGenNet successfully segments nearly all nuclei. These examples demonstrate the model’s robustness in handling complex nuclear morphologies, dense clusters, and heterogeneous tissue types, complementing the quantitative improvements in Table~\ref{table1}.

\begin{figure}[!t]
\includegraphics[width= 3.4in, height=2.25in]{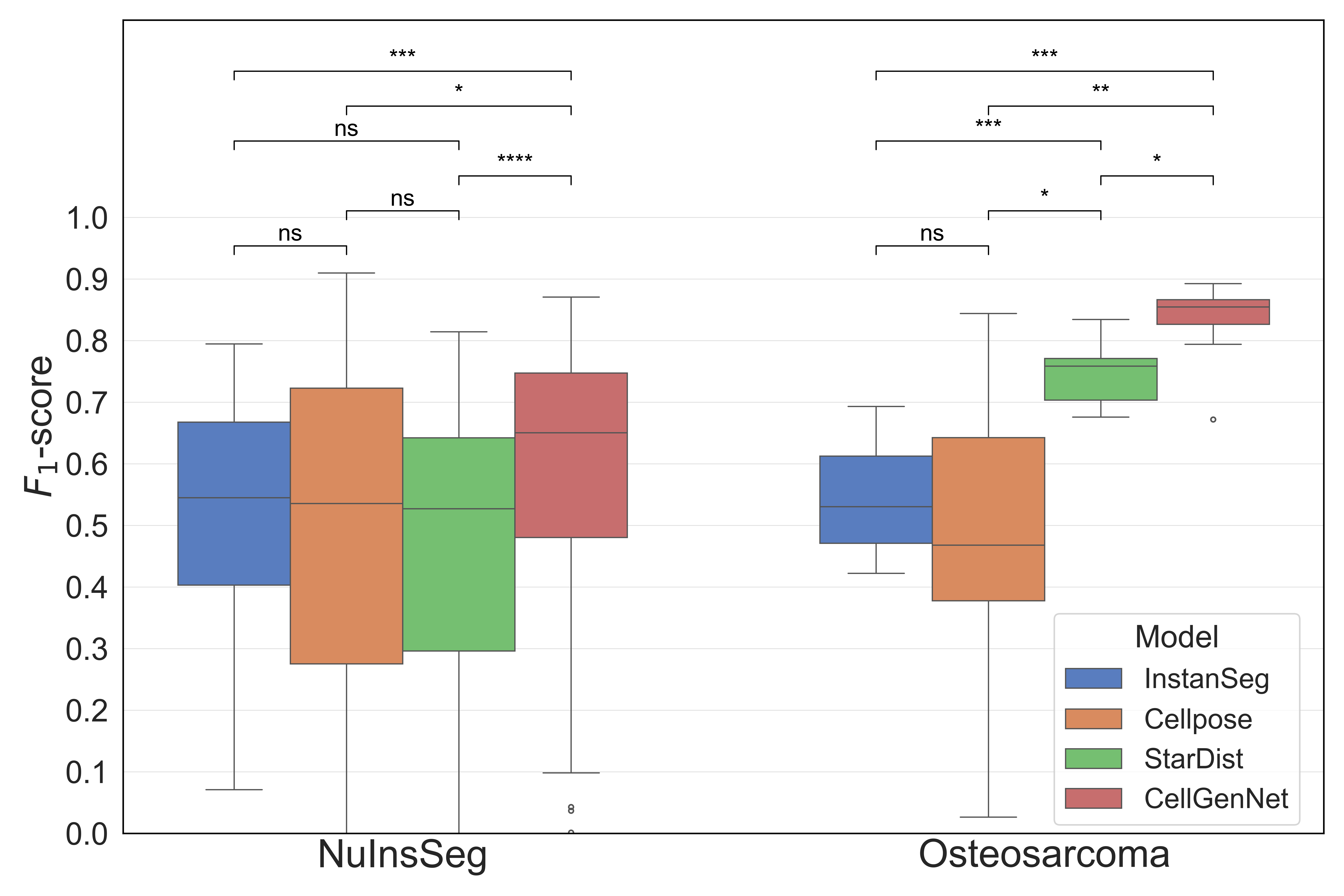}
\caption{Box plots showing the performance of all methods on both datasets and significance levels using a Mann-Whitney U Test. A $p \leq 0.001$ is denoted by three stars "***", a $p \leq 0.01$ is denoted by two stars "**", a $p \leq 0.05$ is denoted by one star "*", and a $p > 0.5$ is denoted by "ns".}
\label{fig3}
\vspace{-5mm}
\end{figure}

Fig.~\ref{fig3} shows box plots of $F_{1}$-score distributions for all images in the Osteosarcoma and NuInsSeg test sets. \textit{CellGenNet} achieves the highest median $F_{1}$ and the tightest interquartile range, indicating superior accuracy and consistency. These improvements are statistically significant across both datasets, highlighting the robustness and generalizability of the model.
\vspace{-2mm}
\section{Conclusion} 

Accurate nuclei segmentation in whole slide images is critical for downstream microscopy analyses but is time-consuming and labor-intensive. We present \textit{CellGenNet}, a knowledge-distillation framework that enables robust segmentation of diverse tissue types without extensive ground-truth annotations. Evaluations on an internal Osteosarcoma dataset and the public NuInsSeg benchmark show that \textit{CellGenNet} outperforms \textit{StarDist}, \textit{Cellpose}, and \textit{InstanSeg}, achieving higher accuracy, better boundary delineation, and consistent detection of complex nuclei. These results demonstrate that combining knowledge distillation with specialized loss functions and regularization produces reliable, generalizable, and scalable models for histopathology and other microscopy applications.

\bibliographystyle{IEEEtran}
\balance
\bibliography{strings,refs}

\end{document}